\documentclass[a4paper, 10pt, conference]{ieeeconf}      

\usepackage{graphics} 
\usepackage{epstopdf}
\usepackage{diagbox}
\usepackage{xcolor}
\usepackage{color,soul}
\definecolor{my_stage1}{RGB}{180,199,231}
\definecolor{my_stage2}{RGB}{255,217,102}
\definecolor{train_shape}{RGB}{143,170,220}
\definecolor{test_shape}{RGB}{244,177,131}
\usepackage{todonotes}
\usepackage{algpseudocode}
\usepackage{algorithm}
\usepackage{epsfig} 
\usepackage{times} 
\usepackage{multirow}
\usepackage{amsmath} 
\usepackage{makecell}
\usepackage{tabularx}
\usepackage{amsfonts}   
\usepackage{ulem}
\usepackage{bm} 
\usepackage{CJK} 
\usepackage{geometry}
\usepackage{graphicx}
\usepackage{booktabs} 
\usepackage{multirow} 
\newcommand{\figref}[1]{Fig.~\ref{#1}}

\usepackage{subfigure}
\usepackage{cite}
\usepackage{gensymb}
\usepackage[colorlinks,
linkcolor=black,
anchorcolor=black,
citecolor=black]
{hyperref}

\usepackage{etoolbox}
\makeatletter
\patchcmd{\@makecaption}
{\scshape}
{}
{}
{}
\makeatother

\IEEEoverridecommandlockouts 
\title{\LARGE \bf
Dual-Actor Fine-Tuning of VLA Models: A Talk-and-Tweak Human-in-the-Loop Approach
}
\author{
    Piaopiao Jin$^{1,*,\dag}$, Qi Wang$^{1,*}$, Guokang Sun$^{1}$, Ziwen Cai$^{1,2}$, Pinjia He$^{2}$, and Yangwei You$^{1}$%
    \thanks{$*$ denotes equal contributions.}%
    \thanks{$\dag$ denotes the corresponding author.}%
    \thanks{$^{1}$ P. Jin, Q. Wang, G. Sun, and Y. You are with Beijing Xiaomi Robot Technology Co., Ltd.
     {\tt\small\{jinpiaopiao, wangqi14, sunguokang, youyangwei1\}@xiaomi.com}}
\thanks{$^{2}$ Z. Cai and P. He are with City University of Hong Kong
     \tt\small\{ziwencai@link.cuhk.edu.cn, hepinjia@cuhk.edu.cn\}}
}

\begin{document}
\maketitle
\thispagestyle{empty}
\pagestyle{empty}

\begin{abstract}
Vision-language-action (VLA) models demonstrate strong generalization in robotic manipulation but face challenges in complex, real-world tasks. While supervised fine-tuning with demonstrations is constrained by data quality, reinforcement learning (RL) offers a promising alternative. We propose a human-in-the-loop dual-actor fine-tuning framework grounded in RL. The framework integrates a primary actor for robust multi-task performance with a refinement actor for latent-space adaptation. Beyond standard physical interventions, we introduce a lightweight talk-and-tweak scheme that converts human corrections into semantically grounded language commands, thereby generating a new dataset for policy learning. In real-world multi-task experiments, our approach achieves 100\% success across three tasks within 101 minutes of online fine-tuning. For long-horizon tasks, it sustains a 50\% success rate over 12 consecutive operations. Furthermore, the framework scales effectively to multi-robot training, achieving up to a 2× improvement in efficiency when using dual robots. The experiment videos are available at \href{https://sites.google.com/view/hil-daft/}{https://sites.google.com/view/hil-daft/}.
\end{abstract}
\vspace{-0.2cm}
\section{INTRODUCTION}
Vision-Language-Action (VLA) models have demonstrated impressive performance across diverse robotic manipulation tasks \cite{pi0,pi05,smolvla,octo_2023,kim24openvla}. Their effectiveness largely arises from large-scale pretraining on demonstration data and alignment with action trajectories. However, despite their generalizable representations, pretrained policies often struggle in complex, real-world scenarios, particularly in long-horizon tasks. Consequently, fine-tuning with domain-specific data is essential for downstream deployment. Supervised Fine-Tuning (SFT) \cite{gunel2020supervised} with human demonstrations remains the dominant approach, but its effectiveness is limited by both the quality and quantity of available data.

Reinforcement Learning (RL)\cite{sutton1998reinforcement} offers a complementary avenue for enhancing policy performance by enabling interaction-based adaptation. Recent advances have demonstrated the effectiveness of RL fine-tuning in pretrained language and vision-language models\cite{lu2025vlarlmasterfulgeneralrobotic,hu2024flareachievingmasterfuladaptive,ouyang2022training}. However, transferring these successes to real-world robotic VLA models remains difficult due to sample inefficiency and safety concerns. To address these challenges, human-in-the-loop strategies that integrate human guidance into online learning have gained traction\cite{luo2024precise,chen2025conrft,luo2024rlifinteractiveimitationlearning}. Among them, Chen et al.\cite{chen2025conrft} proposed a Reinforced Fine-Tuning (RFT) framework that combines offline and online stages, achieving notable sample efficiency in single-task scenarios. Yet, its applicability to multi-task settings and long-horizon manipulations remains unexplored. Moreover, existing human-in-the-loop RL methods typically rely only on physical corrections (e.g., SpaceMouse), lacking a systematic way to transform these interventions into semantically meaningful guidance for policy fine-tuning.

\begin{figure}[t]
    \centering
    \includegraphics[width=0.9\linewidth]{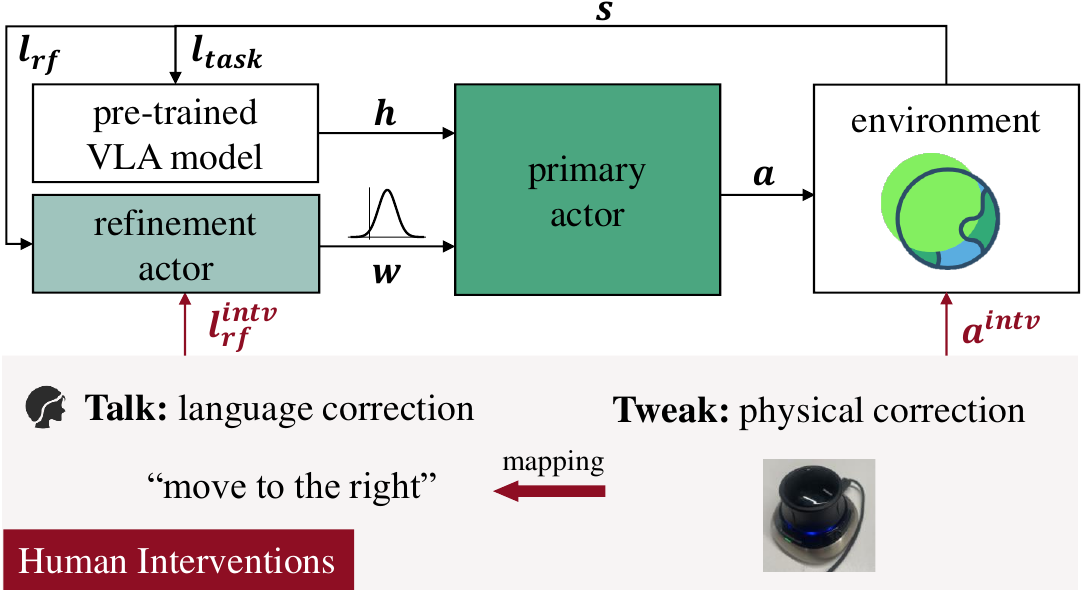}
    \caption{Overview of the human-in-the-loop dual-actor VLA fine-tuning framework. The primary actor generates robust multi-task actions via diffusion, while the refinement actor operates in the latent noise space to provide fine-grained adjustments. Human interventions are integrated through the talk-and-tweak scheme, which translates physical corrections into semantically grounded refinement commands.
}
    \label{framework}
 \vspace{-0.1cm}
 \end{figure}

To address these limitations, we propose a dual-actor human-in-the-loop RL framework (\figref{framework}). The primary actor generates task-general actions, while the refinement actor operates in the latent noise space of the primary policy, providing fine-grained adjustments guided by refinement commands.

To enable efficient dual-actor learning, we introduce a talk-and-tweak intervention scheme. Building on the common practice of correcting actions using a SpaceMouse during rollouts, we extend this paradigm by mapping physical corrections into natural-language refinement commands. For example, a rightward adjustment with the SpaceMouse is verbalized as “move to the right.” The primary actor leverages physical corrections to improve its baseline performance, while the refinement actor learns to associate language commands with targeted adjustments. This design not only preserves general manipulation capabilities but also enables precise, command-driven refinements.

To examine the efficacy of the proposed approach, we study a long-horizon manipulation task where a robot must sequentially place upright, pick up, and assemble bolts scattered on a table (\figref{fig:tasks}). Our approach achieves a 100\% success rate across subtasks after only 101 minutes of online fine-tuning, demonstrating strong sample efficiency. Moreover, it sustains a 50\% completion rate over sequences exceeding 12 steps, underscoring its robustness on long-horizon manipulations. We further extend the framework to multi-robot learning, highlighting its flexibility and potential to improve both sample efficiency and training stability.



Our main contributions are as follows:
\begin{itemize}
\item We propose a novel dual-actor VLA fine-tuning framework that integrates a primary actor for robust multi-task policy generation with a refinement actor operating in the latent noise space to enable fine-grained and controllable action adjustments.
\item We develop a lightweight human-in-the-loop scheme that converts real-time physical corrections (tweak) into semantically grounded language commands (talk), forming a talk-and-tweak dataset. The refinement actor leverages these instructions for interpretable adjustments, while the primary actor improves its baseline policy through direct interventions.
\item We validate the proposed method on real robots, illustrating rapid multi-task fine-tuning with 100\% success within 101 minutes. For long-horizon sequences, it sustains a 50\% success rate over 12 consecutive operations. Moreover, the framework scales seamlessly to multi-robot training, yielding up to a 2× improvement in efficiency when training with dual robots.
\end{itemize}

\section{RELATED WORK}
\subsection{Reinforced Fine-Tuning for Large Models}
Reinforcement Learning (RL) has been widely applied to fine-tune Large Language Models (LLMs) and Vision-Language Models (VLMs) \cite{ouyang2022training, carta2023grounding}. These approaches typically employ on-policy algorithms [15] to adapt pretrained models using human-aligned reward signals. However, they generally require large volumes of interaction data, which constrains their applicability to real-world robotics. To mitigate this limitation, Chen et al.\cite{chen2025conrft} proposed an efficient RL framework for fine-tuning VLA models in physical environments, but their approach was restricted to single-task scenarios. Building on this foundation, our dual-actor fine-tuning framework extends RL-based adaptation to multi-task settings while further enhancing sample efficiency.

\subsection{Human-in-the-Loop Robot Learning}
Human-in-the-loop learning incorporates human input into robot training to improve sample efficiency and ensure safety. This input can take various forms, including demonstrations \cite{kelly2019hg}, interventions \cite{luo2024precise}, preference feedback, and natural language guidance \cite{shi2024yell, Kim2025}.A widely used method in imitation learning is DAgger \cite{hg-dagger}, which iteratively integrated corrective actions from a supervisor. Building on this idea, Shi et al. \cite{shi2024yell} replaced action-level corrections with language feedback for more flexible supervision. Human interventions have also proven effective in RL\cite{luo2024precise}, by guiding exploration, helping agents avoid local optima, and enhancing safety.

Despite these advances, most prior work treats input modalities separately, relying either on low-level motion corrections or high-level language guidance. In contrast, our method unifies these modalities by combining language-based refinement commands (e.g., “move to the right”) with real-time physical corrections. This integrated scheme enables human operators to influence both task objectives and action execution, thereby facilitating more effective and precise policy adaptation.

\subsection{Policy Adaptation}
Policy adaptation refers to the general process of improving an initially learned policy, typically obtained through imitation learning or offline reinforcement learning. Online adaptation has been widely explored in sim-to-real transfer, offline-to-online RL, and imitation-to-RL adaptation. For instance, Ankile et al. \cite{ankile2024imitationrefinementresidual} proposed ResiP, which employed a residual RL policy on top of a frozen behavioral cloning model to reduce the sim-to-real gap. Wagenmaker et al. \cite{wagenmaker2025steering} introduced Diffusion Steering via Reinforcement Learning (DSRL), adapting a behavioral cloning policy by optimizing in the latent noise space of an initial diffusion model.

Building on these works, our approach focuses on offline-to-online RL adaptation for diffusion-based VLA models. We propose a dual-actor architecture that simultaneously optimizes the primary diffusion policy and its latent noise space. Unlike DSRL \cite{wagenmaker2025steering}, we modify the latent noise space guided by natural language commands, resulting in a more sample-efficient and interpretable policy adaptation.


\begin{figure}[t]
    \centering
    \includegraphics[width=0.85\linewidth]{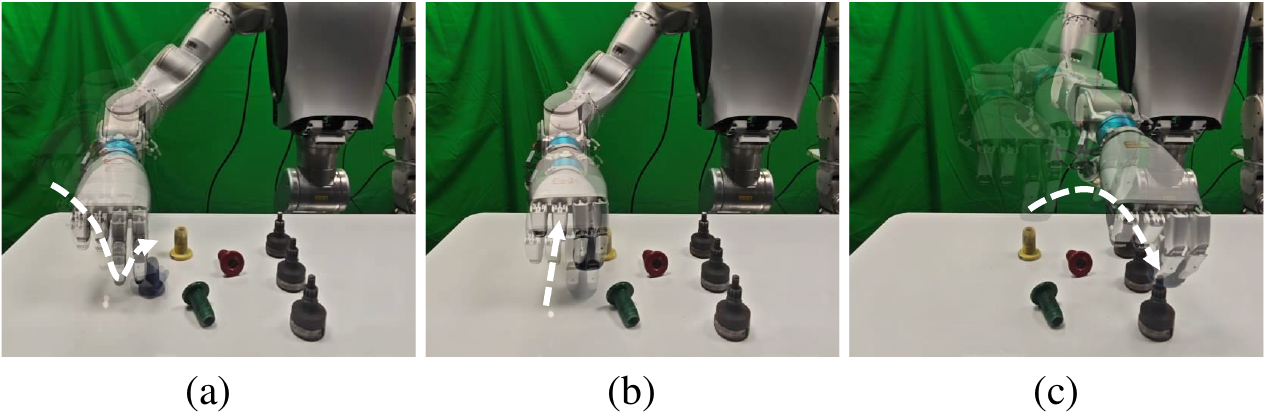}
    \vspace{-0.2cm}
    \caption{Illustration of the three manipulation tasks. (a) Placing the bolt upright, (b) picking up the bolt, and (c) assembling the bolt on the stud.
}
    \label{fig:tasks}
    \vspace{-0.1cm}
\end{figure}

\section{Problem Setup and Preliminaries}
This paper investigates a challenging robotic manipulation task in which a robot must manage bolts scattered on a tabletop, as illustrated in \figref{fig:tasks}. The task consists of three stages: (1) placing the bolt upright by grasping its end, (2) reliably picking up the bolt, and (3) assembling the bolt by inserting it into a partially occluded slot. Each stage requires millimeter-level precision in positioning and alignment. Furthermore, the long-horizon variant of the task demands that the robot sequentially execute these three stages across multiple bolts, chaining them into continuous operations.


We assume access to a pretrained Vision-Language-Action (VLA) model $\mathcal{V}_{\tau_{pre}}$, which encodes RGB observations, proprioceptive states, and task commands. The action head is implemented as a diffusion policy \cite{chi2023diffusion}, where actions are generated by progressively denoising Gaussian noise. Instead of performing multiple denoising steps, we adopt a consistency policy \cite{chen2023boosting, song2023consistencymodels, karras2022elucidating}, which distills the multi-step process into a single-step prediction.

For efficient multi-task generalization and fine-grained adaptation, we adopt a dual-actor learning framework for action generation. In the consistency policy, an action is produced by denoising a Gaussian noise sample:
\begin{equation}\label{eqn:normal_distribution}
w \sim \mathcal{N}(0, K^2I)
\end{equation}
where $K$ is a hyperparameter controlling the initial noise scale. Because different samples of $w$ can lead to different action outputs, we design two complementary modes of action generation.

Let \(\pi_\theta^w\) denote the consistency policy conditioned on the noise \(w\). In the primary mode, $w$ is sampled from Eqn.~\eqref{eqn:normal_distribution}, preserving the original behavior of the policy. In the refinement mode, the noise is instead sampled from a learned distribution:
\vspace{-0.2cm}
\begin{equation}\label{eqn:learned_distribution}
    w  \sim \mathcal{N}(\pi_\phi(\mu|s,l_{rf}),K^2I),
    \vspace{-0.05cm}
\end{equation}
where the mean $\pi_\phi(\mu|s,l_{rf})$ is conditioned on state $s$ and refinement language command $l_{rf}$. 
By maintaining both modes, the dual-actor framework enables straightforward adaptation: when a refinement command is provided, $\pi^w_\phi$ steers the policy toward actions consistent with the instruction; otherwise, the system defaults to the original behavior defined by the primary actor.

\section{Methods}
First, the dual-actor reinforcement learning framework is presented in Sec. IV-A, forming the core of the method. Next, the talk-and-tweak human intervention mechanism is described in Sec. IV-B. Finally, the multi-task learning strategy is depicted in Sec. IV-C.

\subsection{Dual-Actor Reinforcement Learning System}
The robotic manipulation task is formulated as a Markov Decision Process (MDP), $\mathcal{M} = (\mathcal{S}, \mathcal{A}, P, r, \rho, \gamma)$, where $s \in \mathcal{S}$ denotes the state space, $a \in \mathcal{A}$ the action space, $P(s' \mid s, a)$ the transition dynamics, and $\rho(s)$ the initial state distribution. $r(s,a)$ denotes the reward, and $\gamma \in (0,1)$ the discount factor. While the environment executes a single action $a$, our dual-actor framework provides two complementary mechanisms for generating it. In the absence of language-based refinements, the primary actor $\pi^w_\theta$ samples candidate actions by denoising Gaussian noise as defined in Eqn.\eqref{eqn:normal_distribution}. When refinement instructions are provided, $\pi^w_\theta$ instead denoises samples from the learned distribution specified by the refinement actor shown in Eqn.\eqref{eqn:learned_distribution}.

\begin{figure}[t]
    \centering
    \includegraphics[width=0.85\linewidth]{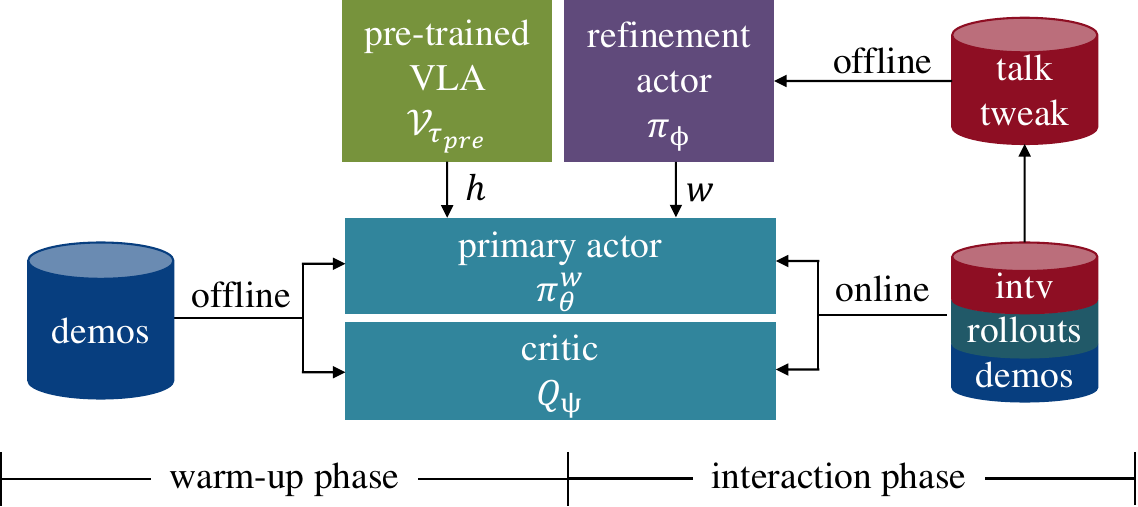}
    \caption{Overview of the two-phase training framework. During the warm-up phase, the primary actor and critic are initialized from offline demonstrations. In the subsequent interaction phase, they are refined using a mixture of offline data, human interventions, and online rollouts, while the refinement actor is trained from the talk-and-tweak dataset.}
    \label{fig:dual-actor}
\end{figure}

The dual-actor framework is optimized through two consecutive training stages: an offline warm-up phase and an online interaction phase (\figref{fig:dual-actor}). During the warm-up phase, the primary policy $\pi^w_\theta$ and the Q-function $Q_\psi(s,a)$ are initialized using demonstration data to establish a stable performance baseline. In the online interaction phase, both $\pi^w_\theta$ and $Q_\psi(s,a)$ are further optimized through real-time environment interactions. At the same time, the refinement policy $\pi_\phi$ is introduced to enable language-conditioned adaptation in the latent space.

During the warm-up phase, we use a pretrained Vision-Language-Action (VLA) model $\mathcal{V}_{\tau_{pre}}$ to encode the task command $l_{\text{task}}$ and state $s$ (RGB images and proprioceptive inputs) into a task embedding:
\begin{equation}\label{eqn:h_formula}
h = \mathcal{V}_{\tau_{pre}}(s, l_\text{task}).
\end{equation}
The primary policy $\pi^w_\theta(h)$ is optimized under a hybrid objective of reward maximization and Behavioral Cloning (BC). On the one hand, the policy is encouraged to maximize the expected Q-value of its actions, and the loss function is defined as
\begin{equation}\label{eqn:q-loss}
\mathcal{L}^{Q}_{\pi^w_{\theta}} = - \mathbb{E}_{s \sim \mathcal{D}} \big[ Q_\psi(s, \pi^w_{\theta}(h)) \big].
\end{equation}
The training buffer is initialized from demonstrations $\mathcal{D} = \mathcal{D}_{\text{demos}}$.
On the other hand, a BC objective guides the policy to mimic expert demonstrations $(s, a^*) \sim \mathcal{D}$:
\begin{equation}
\mathcal{L}^{\text{BC}}_{\pi^w_{\theta}} = \mathbb{E}_{(s,a^*) \sim \mathcal{D}} \big[ || \pi^w_{\theta}(h) - a^* ||^2 \big].
\end{equation}
The overall policy loss integrates these two complementary objectives:
\begin{equation}\label{eqn:primary_loss}
\mathcal{L}_{\pi_{\theta}} = \lambda_1 \mathcal{L}^{\text{BC}}_{\pi_{\theta}} + \lambda_2 \mathcal{L}^{Q}_{\pi_{\theta}},
\end{equation}
where $\lambda_1$ and $\lambda_2$ are trade-off coefficients between the two objectives.

During the warm-up phase, Q-function learning is performed using Calibrated Q-Learning (Cal-QL) \cite{nakamoto2023cal}, following the work of \cite{chen2025conrft}. This approach explicitly mitigates the impact of out-of-distribution actions, thereby stabilizing policy improvement from offline data.

In the online phase, the robot directly interacts with the physical environment for continued learning. The primary actor $\pi^w_\theta(h)$ is optimized with the same loss function as in Eqn.~\eqref{eqn:primary_loss}, while the training data $\mathcal{D}$ is drawn equally from the demonstration buffer $\mathcal{D}_{\text{demos}}$ and the online rollout buffer $\mathcal{D}_{\text{rollouts}}$. During this stage, the weight on the Q-function loss $\lambda_2$ is increased, while the behavioral cloning loss $\lambda_1$ is down-weighted. This scheduling improves training stability while also ensuring effective reward maximization as online experience accumulates. The Q-function $Q_\psi(s, \pi^w_{\theta}(h))$ in this phase is optimized using a standard Bellman loss rather than the Cal-QL variant.

The refinement actor $\pi_\phi$ is trained with offline RL. Conditioned on state $s$ and the refinement command $l_{rf}$, it predicts the mean of the noise distribution $w$:
\begin{equation}
    \mu = \pi_{\phi}(s, l_{rf}).
\end{equation}
The refinement actor encodes RGB images using a ResNet\cite{he2015deepresiduallearningimage} and refinement command using a T5 model\cite{raffel2020exploring}. Each embedding is projected to a common dimensionality through a separate Multi-Layer Perceptron (MLP). These feature vectors are then concatenated and processed by a final MLP to generate the output $\mu$.

The refinement actor is optimized with several objectives.
We first impose a BC loss to guide the policy towards demonstrated behaviors.
\begin{equation}
    \mathcal{L}^{\text{BC}}_{\pi_\phi} = \mathbb{E}_{(s,a^*) \sim \mathcal{D}_{\text{intv}}} \big[ \| \pi_{\theta}^w(h) - a^* \|^2 \big],
\end{equation}
where $\mathcal{D}_{\text{intv}}$ denotes the talk-and-tweak human intervention dataset which will be introduced in Sec. IV-B. Building on this, we enable further refinement by incorporating a Q-function maximization term:
\vspace{-0.1cm}
\begin{equation}
    \mathcal{L}^{Q}_{\pi_\phi} = - \mathbb{E}_{ w  \sim \mathcal{N}(\pi_\phi(\mu|s,l_{rf}),K^2I)} \big[ Q_\psi(s, \pi_{\theta}^w(h)) \big].
\end{equation}
$\pi_{\theta}^w(h)$ denotes the primary policy conditioned on the learned latent variable $w$. To further stabilize training, we introduce a regularization term that constrains the refinement actor to remain consistent with the initial policy when no explicit refinement command is provided. A fixed language instruction “[null]" is used to represent the absence of a command, ensuring consistent model inputs:
\vspace{-0.1cm}
\begin{align}
    \mathcal{L}^{\text{Reg}}_{\pi_\phi} 
    &= \mathbb{E}_{s \sim \mathcal{D}_{\text{intv}}} \Big[ 
        \big\| 
        \pi_{\theta}^{w \sim 
        \mathcal{N}(\pi_\phi(\mu|s,l_{rf}),K^2I)}(h) \nonumber \\
    &\quad - \pi_{\theta}^{w \sim \mathcal{N}(0,K^2I)}(h) 
        \big\|^2 
    \Big].
\end{align}

The overall training objective combines these components:
\vspace{-0.1cm}
\begin{equation}\label{eqn:refinement_loss}
\mathcal{L}_{\pi_{\phi}} = \eta_1 \mathcal{L}^{\text{BC}}_{\pi_{\psi}} + \eta_2 \mathcal{L}^{\text{Q}}_{\pi_{\psi}} \\
+ \eta_3 \mathcal{L}^{\text{Reg}}_{\pi_{\psi}},
\end{equation}
where $\eta_1$, $\eta_2$, $\eta_3$ are are trade-off coefficients between the three objectives.

\begin{figure}[t]
    \centering
    \includegraphics[width=0.75\linewidth]{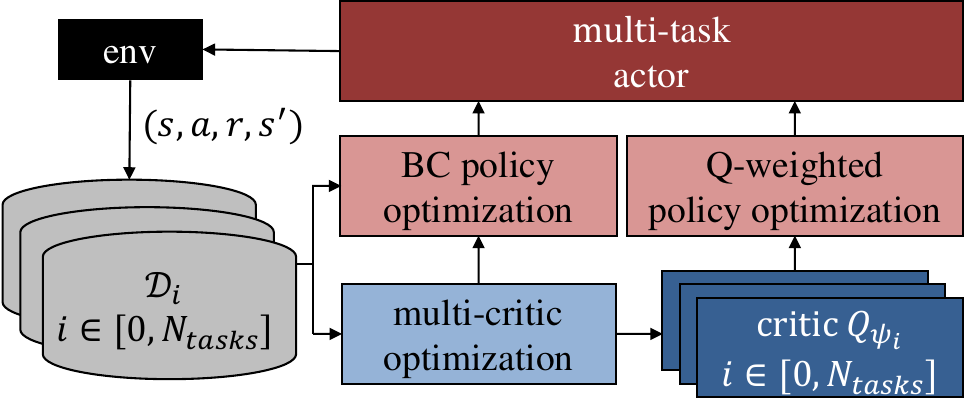}
    \caption{Overview of the multi-task learning architecture. The framework couples a single shared actor with per-task critics, enabling consistent policy learning while maintaining task-specific value functions.
}
    \label{fig:multi-task RL}
\end{figure}

\subsection{Effective Talk-and-Tweak Human Intervention Design}
Human interventions are incorporated into the online learning phase to ensure safe interactions and accelerate policy convergence. Specifically, a human supervisor can directly intervene whenever the agent executes unsafe or suboptimal actions. The resulting action pairs $(s_t, a_t^\text{intv})$ are stored in the intervention dataset $\mathcal{D}_\text{intv}$, which serves as the basis for subsequent policy updates.


Building on this foundation, we introduce a talk-and-tweak dataset generation scheme. In this scheme, each physical correction (tweak) is paired with a corresponding natural language refinement command (talk) via a rule-based mapping function $f: a^{\text{intv}} \mapsto l_{rf}$. Human interventions typically span multiple consecutive time steps. Let $a_{t:t+J-1}^{\text{intv}} = \{a_t^{\text{intv}}, \dots, a_{t+J-1}^{\text{intv}}\}$ denote a sequence of intervened actions, where each $a_t^{\text{intv}} \in \mathbb{R}^7$ encodes the robot action. The first three dimensions represent translational motions $(\Delta x, \Delta y, \Delta z)$, the next three represent rotational motions $(\Delta \text{roll}, \Delta \text{pitch}, \Delta \text{yaw})$, and the last dimension denotes the gripper command. Our language mapping focuses exclusively on the translational components. The cumulative displacement over the time window $J$ is computed as
\begin{equation}
\Delta_t = \sum_{j=0}^{J-1} a_{t+j}^{\text{intv}}[0:3] 
\end{equation}
where $J$ denotes the time window of $J=5$ steps. For each translational axis, if the cumulative displacement exceeds a predefined threshold $\sigma=0.001$m, a corresponding refinement command is generated:

\begin{equation}
\vspace{-0.1cm}
l_{rf_t}^d =
\begin{cases}
\text{``positive $d$ direction''}, & \Delta_t^d > \sigma,\\
\text{``negative $d$ direction''}, & \Delta_t^d < -\sigma,\\
\text{no command}, & |\Delta_t^d| \le \sigma,
\end{cases}
\end{equation}
where $d \in \left({x,y,z}\right)$ denotes the translational axis.

The final refinement command $l_{rf_t}$ is obtained by concatenating the commands across the three translational axes, producing refinement commands such as “move right and forward”. The resulting talk-and-tweak triplet $(s_t, a_t^\text{intv}, l_{{rf}_t})$ replaces the standard intervention pair $(s_t, a_t^\text{intv})$ and is added to the intervention buffer for policy learning. This augmentation transforms raw corrective actions into semantically grounded refinements, enabling the refinement actor to map the 
high-level linguistic guidance to physical actions.


\subsection{Efficient Multi-Task Learning}
We propose a multi-task learning framework that explicitly supports scalable and stable learning across diverse tasks. As illustrated in \figref{fig:multi-task RL}, the architecture employs a shared multi-task actor and task-specific critics $Q_{\psi_i}(s,a)$, where $i$ denotes the task identifier. To be specific, each critic encodes RGB images with a ResNet \cite{he2015deepresiduallearningimage}, proprioceptive inputs with a Multi-Layer Perceptron (MLP), and the action $a$ with another MLP. The resulting features are concatenated and passed through an MLP head to produce the Q-value $Q_{\psi_i}(s,a)$.



To support multi-task training, we extend the task-level replay buffer mechanism proposed by \cite{luo2024precise}. Specifically, for each task \(i\), we maintain three separate buffers: expert demonstrations \(\mathcal{D}^i_{\text{demos}}\), policy-generated rollouts \(\mathcal{D}^i_{\text{rollouts}}\), and human interventions \(\mathcal{D}^i_{\text{intv}}\). During the warm-up phase, \(\mathcal{D}^i_{\text{demos}}\) contains only offline demonstrations. In the subsequent online interaction phase, it is augmented with intervention data from \(\mathcal{D}^i_{\text{intv}}\).  

The primary actor during the interaction phase is updated by uniformly sampling across all tasks and buffer types, i.e.,
$\frac{1}{N}\sum_{i=1}^N \left(\mathcal{D}^i_{\text{demos}} \cup \mathcal{D}^i_{\text{rollouts}}\right)$,
which encourages balanced learning from both demonstrations and autonomous interactions. The refinement actor is trained using the aggregated intervention dataset $\frac{1}{N}\sum_{i=1}^N \mathcal{D}^i_{\text{intv}}$.
In contrast, each task-specific critic is updated exclusively using data from its corresponding task buffers
$\mathcal{D}^i_{\text{demos}} \cup \mathcal{D}^i_{\text{rollouts}}$,
ensuring stable and task-specific value estimation. The detailed procedure is summarized in Algorithm \ref{alg:algorithm}.

\vspace{-0.2cm}

\begin{algorithm}[H]
\caption{Dual-actor Multi-Task Fine-Tuning Algorithm}
\begin{algorithmic}[1]
\Require A pre-trained VLA model $\mathcal{V}_{\tau_{pre}}$, consistency policy head $\pi^w_\theta$, refinement policy $\pi_\phi$, per-task critic models $\{Q_{\psi_i}\}_{i=1}^{N_{tasks}}$. Per-task demonstration datasets $\{\mathcal{D}^i_{\text{demos}}\}_{i=1}^{N_{tasks}}$, each containing 20 demonstrations. Per-task rollout buffers $\{\mathcal{D}^i_{\text{rollouts}}\}_{i=1}^{N_{tasks}}$, intervention buffers $\{\mathcal{D}^i_{\text{intv}}\}_{i=1}^{N_{tasks}}$, and global talk-tweak buffer $\mathcal{D}_{\text{talk-tweak}}$. Batch size $B$.

\State Randomly initialize $\pi^w_\theta$, $\pi_\psi$, and  $\{Q_{\psi_i}\}_{i=1}^{N_{tasks}}$

\Statex \textcolor{gray}{\# I: Offline warm-up phase} 
\For{each offline training step}
    \For{each task $i = 1,\dots,N_{tasks}$}
        \State Sample minibatch $\mathcal{B}^i \sim \mathcal{D}^i_{\text{demos}}$ of size $\tfrac{B}{N_{tasks}}$
        \State Update critic $Q_{\psi_i}$ with $\mathcal{B}^i$
    \EndFor
    \State Update shared consistency policy $\pi^w_\theta$ with $\{\mathcal{B}^i\}_{i=1}^{N_{tasks}}$
\EndFor

\Statex \textcolor{gray}{\# II: Online interaction phase} 
\Statex \textcolor{blue}{Learning Thread:}
\State Wait until each buffer $\mathcal{D}^i_{\text{rollouts}}$ contains at least 100 transitions
\For{each online training step}
    \For{each task $i = 1,\dots,N_{tasks}$}
        \State Sample $\tfrac{B}{2N_{tasks}}$ from $\mathcal{D}^i_{\text{demos}}$ and $\tfrac{B}{2N_{tasks}}$ from $\mathcal{D}^i_{\text{rollouts}}$ to form minibatch $\mathcal{B}^i$
        \State Update critic $Q_{\psi_i}$ with $\mathcal{B}^i$
    \EndFor
    \State Update consistency actor $\pi^w_\theta$ with $\{\mathcal{B}^i\}_{i=1}^{N_{tasks}}$
    \State Update refinement actor $\pi_\phi$ with $\mathcal{D}_{\text{talk-tweak}}$
\EndFor

\Statex \textcolor{blue}{Interaction Thread:}
\For{each task $i = 1,\dots,N_{tasks}$}
    \For{each interaction step}
        \If{no human intervention}
            \State Take action $a_t \sim \pi^w_\theta(s_t, l_{task})$
            \State Store $(s_t, a_t, r_t, s_{t+1})$ in $\mathcal{D}^i_{\text{rollouts}}$
        \Else
            \State Take human-intervened action $a^{\text{intv}}_t$
            \State Store $(s_t, a^{\text{intv}}_t, r_t, s_{t+1})$ in $\mathcal{D}^i_{\text{intv}}$
        \EndIf
    \EndFor
    \State Augment $\mathcal{D}_{\text{talk-tweak}} \gets \mathcal{D}_{\text{talk-tweak}} \cup \mathcal{D}^i_{\text{intv}}$
    \State Augment $\mathcal{D}^i_{\text{demos}} \gets \mathcal{D}^i_{\text{demos}} \cup \mathcal{D}^i_{\text{intv}}$
\EndFor

\end{algorithmic}
\label{alg:algorithm}
\end{algorithm}
\vspace{-0.3cm}

To balance learning progress across multiple tasks, we reweight Q-loss in Eqn.~\eqref{eqn:q-loss}. Instead of a uniform mean,
\begin{equation}
- \frac{1}{N} \sum_{i=1}^N \mathbb{E}_{s \sim \mathcal{D}_i} \big[ Q_{\psi_i}(s, \pi^w_{\theta}(h)) \big], \quad 
\mathcal{D}_i = \mathcal{D}^i_{\text{demos}} \cup \mathcal{D}^i_{\text{rollouts}},
\end{equation}
each task is assigned with a coefficient \(\epsilon_i\) that are inversely proportional to the current Q-value of each task. Tasks with higher Q-values, which indicate stronger performance, are assigned smaller weights, thereby reducing their gradient contribution.Conversely, tasks with lower Q-values receive larger weights, guiding the optimization toward underperforming tasks.
This adaptive weighting scheme mitigates over-optimization of dominant tasks and promotes balanced multi-task learning. Accordingly, the multi-task actor loss is formulated as
\begin{equation}
- \frac{1}{N} \sum_{i=1}^N \epsilon_i \cdot \mathbb{E}_{s \sim \mathcal{D}_i} \big[ Q_{\psi_i}(s, \pi^w_{\theta}(h)) \big],
\vspace{-0.1cm}
\end{equation}
where the task weight \(\epsilon_i\) is defined as

\begin{equation}
\epsilon_i =
\begin{cases}
\epsilon_{max}, & \epsilon_i > \epsilon_{max},\\
\frac{\sum_{i=1}^N \overline{Q}_i}{N\overline{Q}_i+Nc},  & \epsilon_{min}<\epsilon_i \le \epsilon_{max},\\  
\epsilon_{min}, & \epsilon_i \le \epsilon_{min}, 
\end{cases} 
\end{equation}
$\overline{Q}_i = \mathbb{E}_{s \sim \mathcal{D}_i} \big[ Q_{\psi_i}(s, \pi^w_{\theta}(h)) \big] $ denotes the Q-value of task i. $c=0.1$ is a constant that ensures numerical stability for small values of $\overline{Q}$. $\epsilon_{max}=1.2$ and $\epsilon_{min}=0.8$ are used to clip the $\epsilon_i$ into the interval $[\epsilon_{min},\epsilon_{max}]$.

\section{Experiments}
\subsection{Task Design and Experiments Setup}
The primary objective of our experiments is to assess the effectiveness of the proposed fine-tuning method in multi-task learning and long-horizon manipulation. To this end, the experiments are structured around the following research questions (\textbf{RQ}s):

\textbf{RQ1:} How effective is the proposed fine-tuning framework in enabling multi-task learning?

\textbf{RQ2:} How does the proposed dual-actor framework, supported by talk-and-tweak interventions, improve training efficiency and final policy performance?

\textbf{RQ3:} Can the fine-tuning method generalize across different VLA architectures, demonstrating adaptability beyond a specific model design?

\textbf{RQ4:} How robustly does the method maintain performance when scaling to long-horizon tasks?

\textbf{RQ5}: How well does our RL framework scales to multi-robot training and improves training efficiency?

\begin{table*}[t]
\centering
\caption{Comparison of success rates (\%) and episode lengths across different methods}
\vspace{-0.25cm}
\begin{tabularx}{\textwidth}{l|XXXXX|XXXXX}
\hline
\multirow{2}{*}{Task} & \multicolumn{5}{c|}{Success Rate (\%)} & \multicolumn{5}{c}{Episode Length} \\
 & HG-DAgger\cite{hg-dagger} & HIL-ConRFT\cite{chen2025conrft} & DSRL\cite{wagenmaker2025steering} & Ours (w/o dual-actor) & Ours (w/ dual-actor)& HG-DAgger\cite{hg-dagger} & HIL-ConRFT\cite{chen2025conrft} & DSRL\cite{wagenmaker2025steering} & Ours (w/o dual-actor) & Ours (w/ dual-actor) \\
\hline
place the bolt upright  & 28  & 0 & 28  & 88 & \textbf{100} &  41  & - & 37  & 28 & 30 \\
pick up the bolt        & 28  & 0 & 28  & 96 & \textbf{100} &  44  & - & 24  & 32 & 31 \\
assemble the bolt       & 20  & 0 & 16  & 76 & \textbf{100} &  38  & - & 21  & 32 & 31 \\

\hline
Average           & 25.3 & 0  &24.0  & 86.7 & \textbf{100.0} & 41.0  & - & 27.3 & 30.7 & 30.7\\
\hline
\end{tabularx}
\label{tbl:rq1_results}
\vspace{-0.2cm}
\end{table*}

Our framework uses Octo \cite{octo_2023} as the backbone VLA model to extract representations $h$. This model explicitly produce a CLS token $h_{\text{CLS}} \in \mathbb{R}^{B \times D_{\text{CLS}}}$ to encode task representations, where $B$ denotes the batch size and $D$ the head dimension. This token is directly used as $h$. For all tasks, the observation includes two RGB images and the robot’s proprioceptive state. The images are captured from a wrist-mounted camera at 128×128 resolution and a head-mounted camera at 256×256 resolution. The proprioceptive state only includes the gripper status. The action space is defined as a 7-d end-effector delta pose. The first 6 dimensions specify the translational and rotational displacements, and the last dimension represents the binary gripper action for grasping. Data collection and policy execution are performed at a frequency of 10Hz. All experiments are conducted on a custom-built 7 Degree-of-Freedom (DoF) robotic arm developed in-house. The actor processes, responsible for real-time policy execution, run on an NVIDIA Jetson Orin installed on the robot. The learner, which performs offline and online policy updates, is executed on a workstation equipped with an NVIDIA RTX 3090 GPU.

\textbf{Implementation Details:}
During the warm-up phase, policies are initialized with 20 trajectories per task, yielding about 3k state–action pairs. In the subsequent online fine-tuning phase, roughly 15k interaction pairs are collected across the three tasks, corresponding to around 100 trajectories per task. To enhance generalization, both the robot and object positions are uniformly randomized within [-5 cm, 5 cm] along the $x$–$y$ axes. During this phase, human interventions are translated into semantically grounded refinement commands, which account for approximately 15\% of the data and are used to train the refinement policy.The reward function is sparse, assigning a value of 1 for success and 0 otherwise. The loss weights are set as follows: $\lambda_1$ and $\lambda_2$ in Eqn.~\eqref{eqn:primary_loss} are $(1.0,0.1)$ during warm-up and $(0.5,0.5)$ during online interaction, while $\eta_1$, $\eta_2$, and $\eta_3$ in Eqn.~\eqref{eqn:refinement_loss} are $(1.0,0.1,0.1)$.

\textbf{Evaluation Metrics:} For each subtask, including placing the bolt upright, picking up the bolt, and assembling the bolt, a trial is considered successful if the robot completes the task within 50 time steps. Trials that exceed this limit are counted as failures. Success rates are evaluated independently for each subtask. For long-horizon manipulation tasks, full success requires the robot to sequentially complete all three subtasks. During testing, a human operator can issue refinement commands to correct erroneous actions, thereby improving task accuracy and overall performance.

\vspace{-0.1cm}
\subsection{Multi-Task Fine-tuning Capability (RQ1)}
We evaluate the proposed fine-tuning method on multi-task learning against three baselines: HG-Dagger \cite{hg-dagger}, which fine-tunes policies via supervised learning from human corrections; HIL-ConRFT \cite{chen2025conrft}, which applies human-in-the-loop RL with a flat optimization scheme; and DSRL \cite{wagenmaker2025steering}, which refines policies by steering latent noise in diffusion models. For fairness, all methods adopt a two-phase training scheme and consume a comparable amount of interaction data. Results averaged over 25 trials are summarized in TABLE~\ref{tbl:rq1_results}.

As shown in TABLE~\ref{tbl:rq1_results}, our method consistently achieves the highest success rates across all tasks, while also reducing episode length, indicating not only more reliable task completion but also efficient execution.

The HG-Dagger\cite{hg-dagger} method achieves only a 22\% average success rate across the three tasks, primarily due to the limited quality and consistency of the demonstration data. By contrast, our method incorporates reinforcement learning with a Q-loss maximization term, enabling exploration beyond demonstrations and achieving substantially higher performance.
HIL-ConRFT \cite{chen2025conrft} fails completely (0\% success), revealing the weakness of flat optimization in multi-task coordination. Equal data sampling and weighted Q-values in our framework avoid task imbalance, while per-task critics provide more accurate value estimation, leading to stable multi-task learning. DSRL \cite{wagenmaker2025steering} achieves 24\% success by steering latent noise, but its reliance on critics trained solely on offline data limits value accuracy and exploration, degrading fine-tuning performance. An ablation study without the dual-actor structure reaches 86.7\% success, showing the benefits of multi-task balancing. Incorporating the dual-actor scheme further boosts performance to 100\%.

To conclude, these results demonstrate that our method is the only approach capable of achieving reliable 100\% success across diverse manipulation tasks, highlighting the critical role of our design in enabling stable and efficient multi-task learning.



\vspace{-0.1cm}
\subsection{Dual-Actor Policy Learning with Talk-and-Tweak Interventions (RQ2)}
\begin{figure}
    \centering
    \setlength{\abovecaptionskip}{-0.2cm}
    \includegraphics[width=1.0\linewidth]{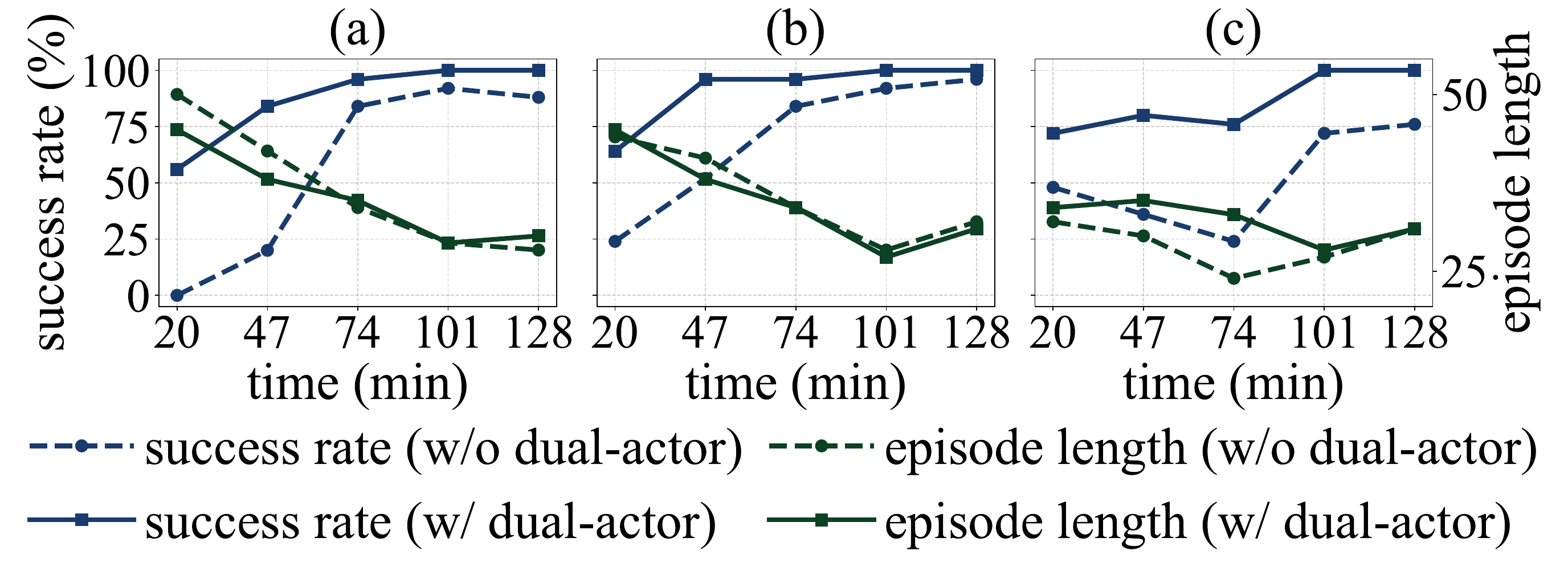}
    \caption{Comparison of success rates and episode lengths across three tasks. Results are shown for the proposed dual-actor method and the ablated single-actor variant over different training durations.}
    \label{fig:rq2}
\end{figure}

This section investigates the effect of the dual-actor learning framework on training stability and efficiency. We compare task success rates and episode length between policies trained w/wo talk-and-tweak intervention scheme. Results averaged over 25 trials are summarized in \figref{fig:rq2}.

As shown in \figref{fig:rq2}, policies trained with the dual-actor framework exhibit substantially higher sample efficiency. After only 20 minutes of interaction, all three tasks achieve an average success rate of 60\%, compared to 30\% under the single-actor baseline. Moreover, the dual-actor framework reaches 100\% success after 101 minutes of fine-tuning, whereas the single-actor counterpart only achieves 86.7\% after 128 minutes. This advantage arises because natural language interventions provide high-level corrective signals that directly steer the policy toward task-relevant behaviors, avoiding the trial-and-error inefficiency of pure reinforcement learning.



We further observe that the single-actor framework exhibits greater instability in multi-task learning. While the first two tasks show gradual improvements, the third task fluctuates significantly. In contrast, the dual-actor framework not only converges more quickly on the earlier tasks but also maintains stable training for the third task. This underscores a major limitation of directly fine-tuning policy parameters with RL, which can introduce instability in multi-task optimization. By refining the latent noise space of the diffusion policy instead, our approach achieves a more stable and efficient fine-tuning process.

In summary, the dual-actor framework strikes a favorable balance between sample efficiency and training stability, demonstrating that latent-space refinement provides an effective policy adaptation mechanism across multiple tasks.


\vspace{-0.1cm}

\begin{table}[h]
\centering
\caption{Success rates (\%) of Octo- and SmolVLA-based models on three tasks}
\vspace{-0.25cm}
\begin{tabular}{lcc}
\toprule
Task & Octo (0.27B)\cite{octo_2023} & SmolVLA (0.6B)\cite{smolvla} \\
\midrule
place the bolt upright       & 100 & 100 \\
pick up the bolt          &  100 & 100  \\
assemble the bolt   & 100 & 100 \\
\midrule
\textbf{Average}   & 100 &  100\\
\bottomrule
\end{tabular}
\label{tbl:rq3_results}
\end{table}
\vspace{-0.2cm}

\subsection{VLA Generalization (RQ3)}
This study examines the generalization of the proposed fine-tuning approach by evaluating its performance on another VLA backbone of SmolVLA \cite{smolvla}. The goal is to determine whether our dual-actor fine-tuning framework can consistently improve multi-task learning performance, independent of the underlying VLA backbone.

\begin{figure*}[t]
    \centering
    \includegraphics[width=\textwidth]{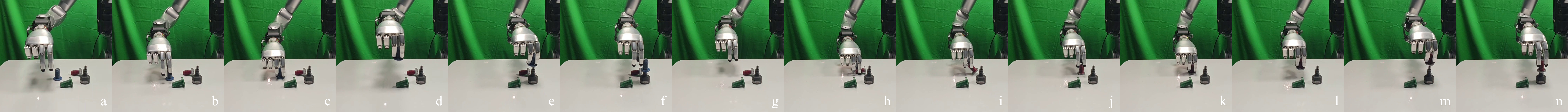}
    \vspace{-0.5cm}
    \caption{Snapshots of the long-horizon manipulation task.}
    \vspace{-0.5cm}
    \label{fig:rq4}
\end{figure*}

As mentioned in Eqn.~\eqref{eqn:h_formula}, our framework employs representations $h$ for policy optimization. Unlike Octo \cite{octo_2023}, which explicitly uses a CLS token for task encoding, SmolVLA \cite{smolvla} lacks such a dedicated token. To address this, we extract task-relevant features directly from the encoder’s attention KV cache. Specifically, given RGB images and task instructions, the encoder computes $K, V \in \mathbb{R}^{B \times T \times H \times D}$, where $B$ is the batch size, $T$ the sequence length, $H$ the number of attention heads, and $D$ the head dimension. A prefix mask $m \in \{0,1\}^{B \times T}$ highlights task-critical tokens. From the final encoder layer, we perform mask-guided average pooling:
\begin{align}
K_{\text{pool}} &= \frac{\sum_{t=1}^T K_{\text{prefix}}}{\sum_{t=1}^T m}, \quad
V_{\text{pool}} = \frac{\sum_{t=1}^T V_{\text{prefix}}}{\sum_{t=1}^T m}.
\end{align}
$K_{\text{prefix}} = K \cdot m$ and $V_{\text{prefix}} = V \cdot m $.  The resulting pooled tensors are flattened and concatenated to form the task embedding $E_{\phi}(s) \in \mathbb{R}^{B \times 2HD} $, which serves as the compact task embedding. By synthesizing visual-language interactions through prefix-guided compression, the embedding $h = E_{\phi}(s)$ provides an efficient and expressive representation for downstream policy optimization tasks.

TABLE~\ref{tbl:rq3_results} illustrates the results averaged across 25 trials; our method consistently achieves strong performance across different VLA backbones, demonstrating its general applicability to diverse pretrained models without backbone-specific modifications.

\subsection{Long-Horizon Task Robustness (RQ4)}

We evaluate the robustness of our dual-actor framework on long-horizon manipulation tasks, where completing a single bolt requires three consecutive steps: placing it upright, picking it up, and assembling it, as showin in \figref{fig:rq4}. To assess scaling with task length, we test consecutive assembly of multiple bolts. The framework achieves 90\% (9/10) success for a single bolt, 60\% (6/10) for two bolts, 60\% (6/10) for three bolts, and 50\% (5/10) for four bolts.

These results demonstrate that our method reliably executes complex, multi-step tasks. While each bolt involves three actions, the framework maintains high performance on single-bolt assembly and generalizes reasonably well to longer sequences. The decline in success with task length mainly stems from error accumulation during assembly, where the partially occluded workspace and blocked slot view make precise alignment challenging.


\subsection{Multi-Robot Scalability and Efficiency (RQ5)}
\begin{figure}
    \centering
    \includegraphics[width=0.75\linewidth]{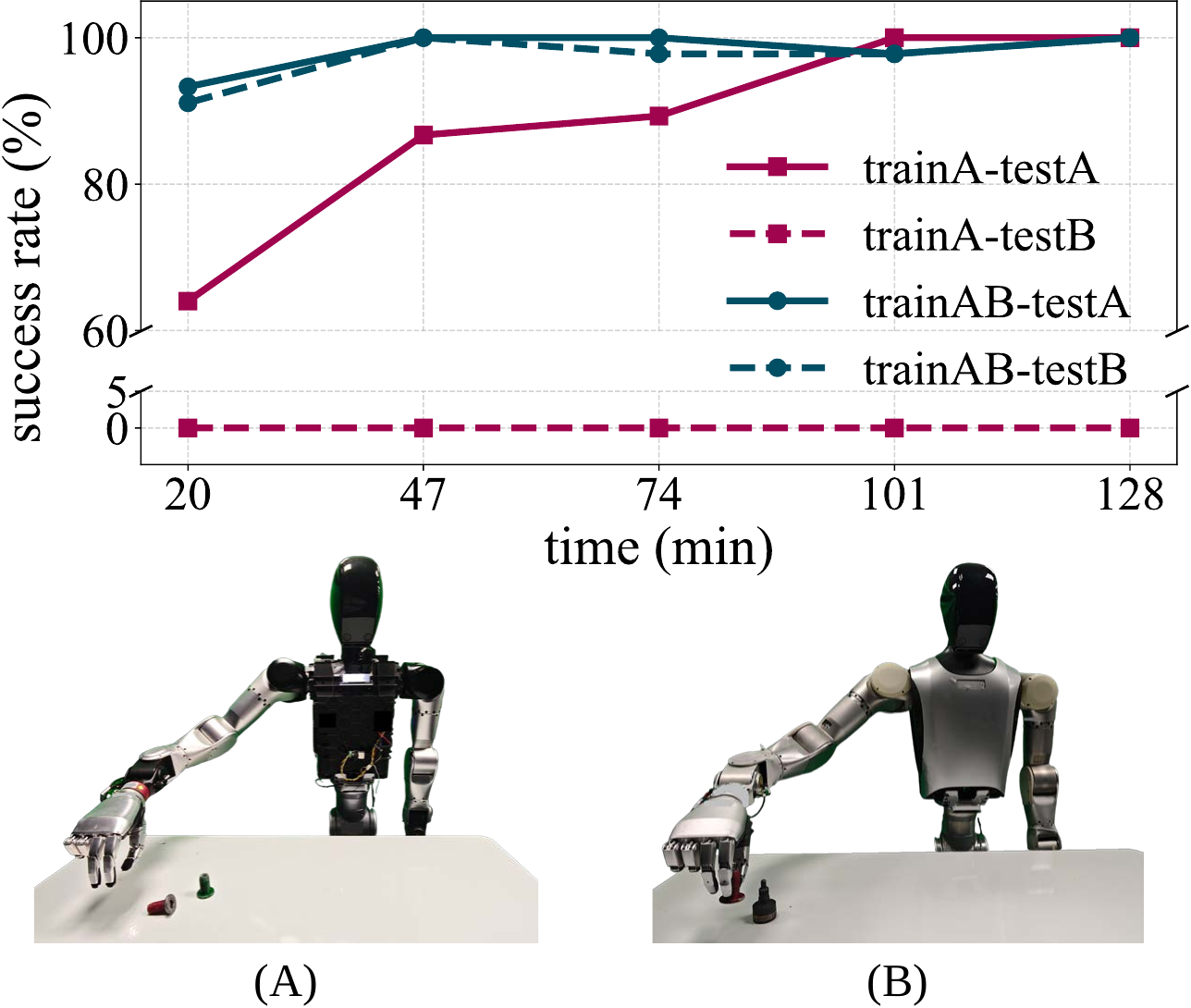}
    \vspace{-0.2cm}
    \caption{Average success rates across three tasks under different training and deployment settings. The first model is trained solely on robot A and evaluated on both robot A (trainA–testA) and robot B (trainA–testB). The second model is trained jointly on data from both robots (trainAB).}
    \label{fig:rq5}
\end{figure}
This experiment evaluates the scalability of our dual-actor fine-tuning framework in a multi-robot training setting. We adopt a centralized learner with decentralized actors: a single learner (server) maintains and updates the policy, while two robotic arms (actors) interact with the environment in parallel. Each actor collects trajectories and transmits them to the learner, which updates the policy and synchronizes the improved version back to both actors.

The warm-up phase follows prior practice by collecting offline demonstration data from a single robot (A). The online phase is then resumed with two robots (A+B). We assess the impact of multi-robot training on both training efficiency and task performance. Results averaged over 25 trials are summarized in \figref{fig:rq5}. We observe that the model trained solely on robot A fails to transfer directly to robot B (100\% $\rightarrow$ 0\%), primarily due to hardware installation and background differences. In contrast, the model co-trained on robots A and B achieves higher sample efficiency, reaching nearly 100\% success after just 47 minutes of fine-tuning, 2× faster than single-robot training. It also maintains consistent and coordinated behavior across agents. These results underscore the potential of our framework for large-scale real-world deployments, where parallel learning substantially reduces training time and improves sample efficiency. Moreover, the ability to sustain high performance across diverse robotic platforms further demonstrates the robustness of our approach.



\section{Conclusions and Limitations}
In this work, we introduce a human-in-the-loop dual-actor fine-tuning framework for vision-language-action models. The primary actor ensures robust multi-task generalization, while the refinement actor leverages language-guided corrections to perform fine-grained adjustments in the latent space. Experiments on real-world multi-task and long-horizon manipulation tasks show that our approach improves training efficiency, success rates, and stability compared to existing fine-tuning methods. Moreover, the framework can be extended to multi-robot setups, enabling parallelized learning and enhanced sample efficiency.

Despite these advantages, our method has some limitations. Performance can degrade in long-horizon sequences due to error accumulation, particularly in occluded or visually challenging environments. Additionally, the reliance on human interventions for refinement introduces a dependency on supervision quality, which may affect scalability in fully autonomous deployments. Future work will explore more autonomous refinement strategies and improved robustness for complex, extended-horizon tasks.

\normalem
\bibliographystyle{IEEEtran}
\bibliography{paper}

\begin{thebibliography}{10}
\providecommand{\url}[1]{#1}
\csname url@samestyle\endcsname
\providecommand{\newblock}{\relax}
\providecommand{\bibinfo}[2]{#2}
\providecommand{\BIBentrySTDinterwordspacing}{\spaceskip=0pt\relax}
\providecommand{\BIBentryALTinterwordstretchfactor}{4}
\providecommand{\BIBentryALTinterwordspacing}{\spaceskip=\fontdimen2\font plus
\BIBentryALTinterwordstretchfactor\fontdimen3\font minus \fontdimen4\font\relax}
\providecommand{\BIBforeignlanguage}[2]{{%
\expandafter\ifx\csname l@#1\endcsname\relax
\typeout{** WARNING: IEEEtran.bst: No hyphenation pattern has been}%
\typeout{** loaded for the language `#1'. Using the pattern for}%
\typeout{** the default language instead.}%
\else
\language=\csname l@#1\endcsname
\fi
#2}}
\providecommand{\BIBdecl}{\relax}
\BIBdecl

\bibitem{pi0}
P.~Intelligence, ``$\pi_0$: A vision-language-action flow model for general robot control,'' 2024.

\bibitem{pi05}
\BIBentryALTinterwordspacing
{Physical Intelligence}, ``$\pi_{0.5}$: a vision-language-action model with open-world generalization,'' 2025. [Online]. Available: \url{https://arxiv.org/abs/2504.16054}
\BIBentrySTDinterwordspacing

\bibitem{smolvla}
\BIBentryALTinterwordspacing
A.~Marafioti, O.~Zohar, M.~Farré, M.~Noyan, E.~Bakouch, P.~Cuenca, C.~Zakka, L.~B. Allal, A.~Lozhkov, N.~Tazi, V.~Srivastav, J.~Lochner, H.~Larcher, M.~Morlon, L.~Tunstall, L.~von Werra, and T.~Wolf, ``Smolvlm: Redefining small and efficient multimodal models,'' 2025. [Online]. Available: \url{https://arxiv.org/abs/2504.05299}
\BIBentrySTDinterwordspacing

\bibitem{octo_2023}
{Octo Model Team}, ``Octo: An open-source generalist robot policy,'' in \emph{Proceedings of Robotics: Science and Systems}, Delft, Netherlands, 2024.

\bibitem{kim24openvla}
M.~Kim, K.~Pertsch, S.~Karamcheti, T.~Xiao, A.~Balakrishna, S.~Nair, R.~Rafailov, E.~Foster, G.~Lam, P.~Sanketi, Q.~Vuong, T.~Kollar, B.~Burchfiel, R.~Tedrake, D.~Sadigh, S.~Levine, P.~Liang, and C.~Finn, ``Openvla: An open-source vision-language-action model,'' \emph{arXiv preprint arXiv:2406.09246}, 2024.

\bibitem{gunel2020supervised}
B.~Gunel, J.~Du, A.~Conneau, and V.~Stoyanov, ``Supervised contrastive learning for pre-trained language model fine-tuning,'' \emph{arXiv preprint arXiv:2011.01403}, 2020.

\bibitem{sutton1998reinforcement}
R.~S. Sutton, A.~G. Barto \emph{et~al.}, \emph{Reinforcement learning: An introduction}.\hskip 1em plus 0.5em minus 0.4em\relax MIT press Cambridge, 1998, vol.~1, no.~1.

\bibitem{lu2025vlarlmasterfulgeneralrobotic}
G.~Lu, W.~Guo, C.~Zhang, Y.~Zhou, H.~Jiang, Z.~Gao, Y.~Tang, and Z.~Wang, ``Vla-rl: Towards masterful and general robotic manipulation with scalable reinforcement learning,'' 2025.

\bibitem{hu2024flareachievingmasterfuladaptive}
\BIBentryALTinterwordspacing
J.~Hu, R.~Hendrix, A.~Farhadi, A.~Kembhavi, R.~Martin-Martin, P.~Stone, K.-H. Zeng, and K.~Ehsani, ``Flare: Achieving masterful and adaptive robot policies with large-scale reinforcement learning fine-tuning,'' 2024. [Online]. Available: \url{https://arxiv.org/abs/2409.16578}
\BIBentrySTDinterwordspacing

\bibitem{ouyang2022training}
L.~Ouyang, J.~Wu, X.~Jiang, D.~Almeida, C.~Wainwright, P.~Mishkin, C.~Zhang, S.~Agarwal, K.~Slama, A.~Ray \emph{et~al.}, ``Training language models to follow instructions with human feedback,'' \emph{Advances in neural information processing systems}, vol.~35, pp. 27\,730--27\,744, 2022.

\bibitem{luo2024precise}
J.~Luo, C.~Xu, J.~Wu, and S.~Levine, ``Precise and dexterous robotic manipulation via human-in-the-loop reinforcement learning,'' \emph{arXiv preprint arXiv:2410.21845}, 2024.

\bibitem{chen2025conrft}
Y.~Chen, S.~Tian, S.~Liu, Y.~Zhou, H.~Li, and D.~Zhao, ``Conrft: A reinforced fine-tuning method for vla models via consistency policy,'' \emph{arXiv preprint arXiv:2502.05450}, 2025.

\bibitem{luo2024rlifinteractiveimitationlearning}
\BIBentryALTinterwordspacing
J.~Luo, P.~Dong, Y.~Zhai, Y.~Ma, and S.~Levine, ``Rlif: Interactive imitation learning as reinforcement learning,'' 2024. [Online]. Available: \url{https://arxiv.org/abs/2311.12996}
\BIBentrySTDinterwordspacing

\bibitem{carta2023grounding}
T.~Carta, C.~Romac, T.~Wolf, S.~Lamprier, O.~Sigaud, and P.-Y. Oudeyer, ``Grounding large language models in interactive environments with online reinforcement learning,'' in \emph{International Conference on Machine Learning}.\hskip 1em plus 0.5em minus 0.4em\relax PMLR, 2023, pp. 3676--3713.

\bibitem{kelly2019hg}
M.~Kelly, C.~Sidrane, K.~Driggs-Campbell, and M.~J. Kochenderfer, ``Hg-dagger: Interactive imitation learning with human experts,'' in \emph{2019 International Conference on Robotics and Automation (ICRA)}.\hskip 1em plus 0.5em minus 0.4em\relax IEEE, 2019, pp. 8077--8083.

\bibitem{shi2024yell}
L.~X. Shi, Z.~Hu, T.~Z. Zhao, A.~Sharma, K.~Pertsch, J.~Luo, S.~Levine, and C.~Finn, ``Yell at your robot: Improving on-the-fly from language corrections,'' \emph{arXiv preprint arXiv: 2403.12910}, 2024.

\bibitem{Kim2025}
J.~W.~B. Kim, J.-T. Chen, P.~Hansen, L.~X. Shi, A.~Goldenberg, S.~Schmidgall, P.~M. Scheikl, A.~Deguet, B.~M. White, D.~R. Tsai, R.~J. Cha, J.~Jopling, C.~Finn, and A.~Krieger, ``Srt-h: A hierarchical framework for autonomous surgery via language-conditioned imitation learning,'' \emph{Science Robotics}, vol.~10, no. 104, p. eadt5254, 2025.

\bibitem{hg-dagger}
\BIBentryALTinterwordspacing
M.~Kelly, C.~Sidrane, K.~R. Driggs{-}Campbell, and M.~J. Kochenderfer, ``Hg-dagger: Interactive imitation learning with human experts,'' \emph{CoRR}, vol. abs/1810.02890, 2018. [Online]. Available: \url{http://arxiv.org/abs/1810.02890}
\BIBentrySTDinterwordspacing

\bibitem{ankile2024imitationrefinementresidual}
\BIBentryALTinterwordspacing
L.~Ankile, A.~Simeonov, I.~Shenfeld, M.~Torne, and P.~Agrawal, ``From imitation to refinement -- residual rl for precise assembly,'' 2024. [Online]. Available: \url{https://arxiv.org/abs/2407.16677}
\BIBentrySTDinterwordspacing

\bibitem{wagenmaker2025steering}
A.~Wagenmaker, M.~Nakamoto, Y.~Zhang, S.~Park, W.~Yagoub, A.~Nagabandi, A.~Gupta, and S.~Levine, ``Steering your diffusion policy with latent space reinforcement learning,'' \emph{Conference on Robot Learning (CoRL)}, 2025.

\bibitem{chi2023diffusion}
C.~Chi, Z.~Xu, S.~Feng, E.~Cousineau, Y.~Du, B.~Burchfiel, R.~Tedrake, and S.~Song, ``Diffusion policy: Visuomotor policy learning via action diffusion,'' \emph{The International Journal of Robotics Research}, p. 02783649241273668, 2023.

\bibitem{chen2023boosting}
Y.~Chen, H.~Li, and D.~Zhao, ``Boosting continuous control with consistency policy,'' \emph{arXiv preprint arXiv:2310.06343}, 2023.

\bibitem{song2023consistencymodels}
\BIBentryALTinterwordspacing
Y.~Song, P.~Dhariwal, M.~Chen, and I.~Sutskever, ``Consistency models,'' 2023. [Online]. Available: \url{https://arxiv.org/abs/2303.01469}
\BIBentrySTDinterwordspacing

\bibitem{karras2022elucidating}
T.~Karras, M.~Aittala, T.~Aila, and S.~Laine, ``Elucidating the design space of diffusion-based generative models,'' \emph{Advances in neural information processing systems}, vol.~35, pp. 26\,565--26\,577, 2022.

\bibitem{nakamoto2023cal}
M.~Nakamoto, S.~Zhai, A.~Singh, M.~Sobol~Mark, Y.~Ma, C.~Finn, A.~Kumar, and S.~Levine, ``Cal-ql: Calibrated offline rl pre-training for efficient online fine-tuning,'' \emph{Advances in Neural Information Processing Systems}, vol.~36, pp. 62\,244--62\,269, 2023.

\bibitem{he2015deepresiduallearningimage}
\BIBentryALTinterwordspacing
K.~He, X.~Zhang, S.~Ren, and J.~Sun, ``Deep residual learning for image recognition,'' 2015. [Online]. Available: \url{https://arxiv.org/abs/1512.03385}
\BIBentrySTDinterwordspacing

\bibitem{raffel2020exploring}
C.~Raffel, N.~Shazeer, A.~Roberts, K.~Lee, S.~Narang, M.~Matena, Y.~Zhou, W.~Li, and P.~J. Liu, ``Exploring the limits of transfer learning with a unified text-to-text transformer,'' \emph{Journal of machine learning research}, vol.~21, no. 140, pp. 1--67, 2020.

\end{thebibliography}

\end{document}